\documentclass{isprs} 
\usepackage{subfigure}
\usepackage{setspace}
\usepackage{geometry} 
\usepackage{epstopdf}
\usepackage{natbib}
\usepackage{amsmath}
\usepackage{array}
\usepackage{booktabs}
\usepackage[labelsep=period]{caption}  
\usepackage[british]{babel} 
\usepackage[hang]{footmisc}

\begin{document}

\title{A Critical Synthesis of Uncertainty Quantification\\ and Foundation Models for Semantic Segmentation}
\date{}


\author{Steven Landgraf \thanks{Corresponding Authors} , Joceline Hinz\textsuperscript{*}, Markus Ulrich}

\address{Institute of Photogrammetry and Remote Sensing (IPF),\\ Karlsruhe Institute of Technology (KIT), Germany,\\ (steven.landgraf, joceline.hinz, markus.ulrich)@kit.edu}

\abstract{
Foundation models are increasingly breaking what seemed to be impossible not long ago by enabling unprecedented accuracy and cross-domain generalization. Yet their lack of interpretability, tendency to be overconfident, and sensitivity to real-world domain shifts pose critical challenges for safety- and mission-critical applications. Uncertainty quantification (UQ) offers a principled way to address these issues, but its integration into segmentation foundation models has yet to be explored. In this paper we present the first systematic evaluation of UQ methods applied to a foundation model for semantic segmentation. We fine-tune a lightweight DPT decoder on top of the pretrained SAM2 encoder to establish a simple yet competitive baseline and benchmark four representative UQ approaches -- Monte Carlo Dropout, Deep Sub-Ensemble, Test-Time Augmentation, and Evidential Deep Learning -- across Cityscapes, NYUv2, and two challenging out-of-domain settings. Our analysis compares segmentation accuracy, calibration, uncertainty quality, and inference time, revealing clear trade-offs between predictive performance, reliability, and computational cost. These results highlight both the promise and the current limitations of uncertainty-aware foundation models, pointing to the need for future work that jointly optimizes accuracy, robustness, and efficiency for real-world deployment.
}

\keywords{Uncertainty Quantification, Reliability, Robustness, Foundation Models, Semantic Segmentation}

\maketitle


\section{Introduction}\label{INTRODUCTION}
\sloppy
Semantic segmentation is a foundational machine vision task involving assigning class labels to every pixel in an image \citep{mo2022review}. Recently, segmentation models have evolved rapidly from convolutional neural networks \citep{li2021survey} and vision transformers \citep{han2022survey} to foundation models trained on internet-scale data \citep{zhou2024image}. These models promise unprecedented accuracy and generalization, enabling zero-shot capabilities across diverse domains and marking a paradigm shift in semantic segmentation.

However, even the most potent neural networks remain subject to critical limitations \citep{awais2025foundation}. Foundation models in particular can act as black boxes, lacking interpretability \citep{gawlikowski2023survey}, failing to recognize out-of-domain inputs \citep{ovadia2019can}, producing systematically overconfident outputs \citep{wilson2020bayesian}, or being sensitive to adversarial perturbations \citep{rawat2017harnessing}. It goes without saying that these issues are particularly harmful in safety- and mission-critical applications, where segmentation errors can propagate to downstream decisions with severe consequences. 

Incorporating uncertainty quantification (UQ) into deep neural networks directly addresses many of the challenges outlined above. Reliable uncertainty estimates not only mitigate overconfidence and sensitivity to domain shifts but also enhance interpretability by indicating the model's confidence and highlighting areas of potential error, thereby improving the deployability deep learning-based systems. For instance, in safety-critical applications like autonomous driving \citep{mcallister2017concrete} or medical imaging \citep{nair2020exploring}, uncertainty-aware models can flag unreliable predictions to trigger human intervention or additional verification steps, reducing the risk of catastrophic failures.

Despite these advantages of UQ, the integration of UQ into foundation models for semantic segmentation remains underexplored. By systematically evaluating how different UQ methods interact with these powerful models, we offer not only improved reliability and explainability but also bridge the gap between state-of-the-art research and real-world deployment. Our contributions can be summarized as follows:
\begin{enumerate}
    \itemsep0em
    \item We present the first systematic study of UQ methods applied to foundation models for semantic segmentation, evaluating Monte Carlo Dropout (MCD), Deep Sub-Ensemble (DSE), Test-Time Augmentation (TTA), and Evidential Deep Learning (EDL).
    \item We establish a simple yet competitive baseline by fine-tuning a DPT head on top of the pretrained SAM2 encoder, and benchmark its in-domain and out-of-domain performance across multiple datasets.
    \item We provide a comprehensive comparison between segmentation performance, calibration, uncertainty quality, and inference time for all methods, highlighting trade-offs between accuracy, reliability, and computational cost. 
\end{enumerate}

\section{Related Work}\label{RELATED_WORK}
Our work lies at the intersection of semantic segmentation and UQ, focusing on adapting foundation models like the Segment Anything Model (SAM) \citep{kirillov2023segment}. We first review advancements in segmentation, from CNNs to vision transformers and foundation models. We then discuss key deep learning UQ methods, including recent efforts to tailor them for foundation models. Finally, we highlight the research gap addressed by this paper.

\subsection{Semantic Segmentation}
\textbf{Convolutional Neural Networks (CNNs).} The foundational work in deep learning-based segmentation began with fully convolutional networks, which enable end-to-end pixel-wise classification \citep{long2015fully}. Subsequent CNN methods introduced richer context and multi-scale features. For instance, DeepLab established dilated convolutions and atrous spatial pyramid pooling to capture image context at multiple scales \citep{chen2017deeplab}. Similarly, encoder-decoder networks like U-Net used symmetric contracting and expanding paths with skip connections to recover even fine details \citep{ronneberger2015u}. Together, these CNN-based approaches represented the state of the art for many years and have only recently been challenged by attention-based architectures.

\textbf{Vision Transformers (ViTs).} Transformer architecture have been the de-facto standard for natural language processing for a long time \citep{vaswani2017attention}. More recently, they have also been brought into the vision domain \citep{dosovitskiy2020image}. In the context of semantic segmentation, SegFormer stands out with its hierarchical transformer encoder and a lightweight multilayer perceptron decoder, achieving impressive results and high efficiency \citep{xie2021segformer}. Mask2Former further unified semantic, instance and panoptic segmentation with a mask-classification transformer, setting new state-of-the-art results \citep{cheng2022masked}. Beyond these task-specific advances, the field has shifted toward foundation models trained on internet-scale data. For example, SAM was trained with over 1B masks on 11M images \citep{kirillov2023segment}. As noted in a recent study \citep{zhou2024image}, adapting these foundation models can yield superior segmentation performance and entirely new capabilities in terms of zero/few-shot segmentation, interactive prompting, or cross-domain generalization that were unseen until now. 

\subsection{Uncertainty Quantification}
\textbf{Overview.} A variety of techniques have been proposed to capture predictive uncertainty in deep learning models \citep{mackay1992practical,gal2016dropout,lakshminarayanan2017simple,valdenegro2023sub,van2020uncertainty,liu2020simple,mukhoti2023deep,amini2020deep}. The most popular methods remain sampling-based due to their ease of use and effectiveness: for instance, MCD \citep{gal2016dropout} treats dropout \citep{srivastava2014dropout} as a Bayesian approximation. Similarly, Deep Ensembles \citep{lakshminarayanan2017simple} consist of multiple, individually trained models to obtain state-of-the-art uncertainty results. Unfortunately, these approaches require multiple forward passes and sometimes more training time, which incurs high computational cost. To mitigate this, a number of deterministic, more efficient, methods have been proposed \citep{van2020uncertainty,liu2020simple,mukhoti2023deep,landgraf2024dudes,landgraf2025efficient}. Likewise, EDL enables efficient uncertainty estimation by training the model to infer the parameters of a probability distribution with a single prediction \citep{amini2020deep}.

\textbf{Uncertainty-aware Foundation Models.} Being able to estimate the uncertainty of the output of large foundation models is an emerging research field. Recently, \citet{landgraf2025critical} studied the synthesis of UQ methods and foundation models in monocular depth estimation and explicitly note that extending this work to other tasks, such as semantic segmentation, is an open opportunity. A few early methods have begun to address UQ for SAM in medical imaging \citep{jiang2024uncertainty,deng2023sam,zhang2023segment}. Beyond, USAM proposes a post-hoc training method for additional multilayer perceptrons to estimate the expected uncertainty \citep{kaiser2025uncertainsam}. \citet{liu2024uncertainty} introduce SUM that quantifies uncertainty in SAM-generated pseudo-labels and uses it to enable uncertainty-aware fine-tuning of the model.

\textbf{Research Gap.} While all of these works have made progress toward uncertainty-aware foundation models in semantic segmentation with tailor-made approaches, they also reveal a critical gap: There has yet to be a systematic study of existing UQ methods in combination with SAM, despite its status as a de-facto foundation model for large-scale segmentation. 

\section{Methodology}\label{METHODOLOGY}
In this section, we describe how we derive our baseline model from the Segment Anything Model 2 (SAM2) foundation model and how we combine four UQ techniques with it.

\subsection{Baseline Model}
\begin{figure}[t!]
\begin{center}
    \includegraphics[width=1.0\columnwidth]{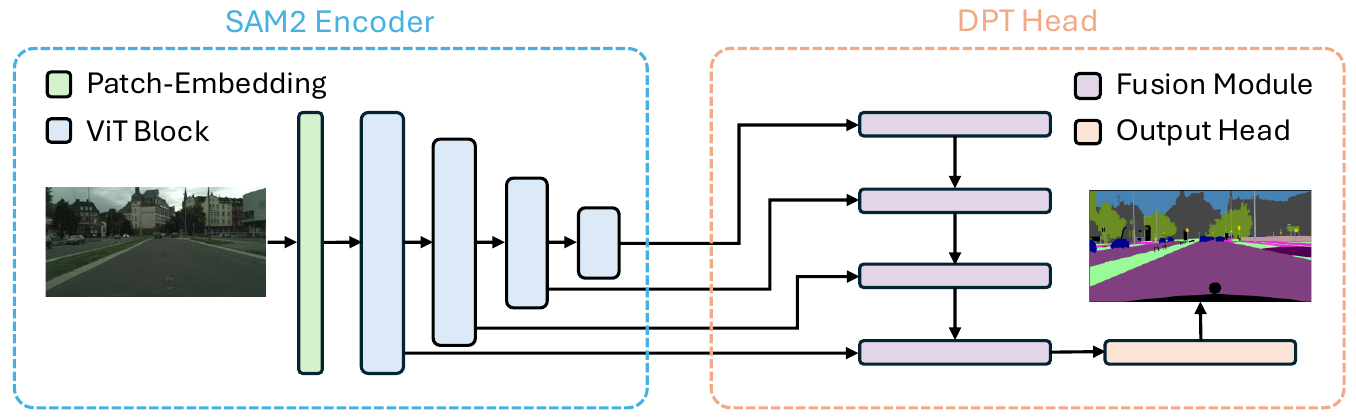}
    \caption{Schematic illustration of the used model architecture.}
\label{fig: architecture}
\end{center}
\end{figure}

We employ a hybrid architecture that combines the image encoder of the SAM2 \citep{ravi2024sam} with the DPT decoder \citep{ranftl2021vision} for semantic segmentation (see Fig.~\ref{fig: architecture}). The motivation for this design is twofold: First, by leveraging the SAM2 encoder, which builds on the hierarchical Vision Transformer Hiera \citep{ryali2023hiera} pretrained with masked autoencoding \citep{he2022masked}, we directly harness state-of-the-art foundation model knowledge in the form of robust, multi-scale image representations. Second, coupling this encoder with the DPT decoder allows us to rely on a simple and widely adopted architecture for dense prediction, ensuring interpretability, comparability, and computational efficiency. The SAM2 encoder partitions the image into patches, projects them into tokens with positional embeddings, and processes them through four hierarchical transformer stages, producing multi-scale feature maps. These are fused by the DPT decoder through RefineNet-based \citep{lin2017refinenet} fusion modules that progressively upsample and align resolution, before passing them to a lightweight segmentation head. The head simply applies a 3x3-convolution, injects a dropout layer, followed by a linear projection, and lastly uses bilinear upsampling to restore the original image resolution.
Training the model is equally simple as its architectural layout: we simply fine-tune the model with the regular cross-entropy loss
\begin{equation}\label{eq: ce}
    \mathcal{L}_{\text{CE}} = -\frac{1}{N}\sum^N_{n=1}\sum^C_{c=1}y_{n,c}\log (p(\hat{y}_{n,c})\enspace,
\end{equation}
where $\mathcal{L}_{\text{CE}}$ represents the loss for a single image, $N$ is the number of pixels, $C$ is the number of classes, $y_{n,c}$ is the one-hot encoded ground truth label, and $p(\hat{y}_{n,c})$ is the predicted softmax pseudo-probability. Overall, our intent was to maximize the benefits of foundation model pretraining while keeping the rest of the architecture as simple as possible.

\subsection{Uncertainty Quantification}
\begin{figure*}[t]
\begin{center}
		\includegraphics[width=1.0\linewidth]{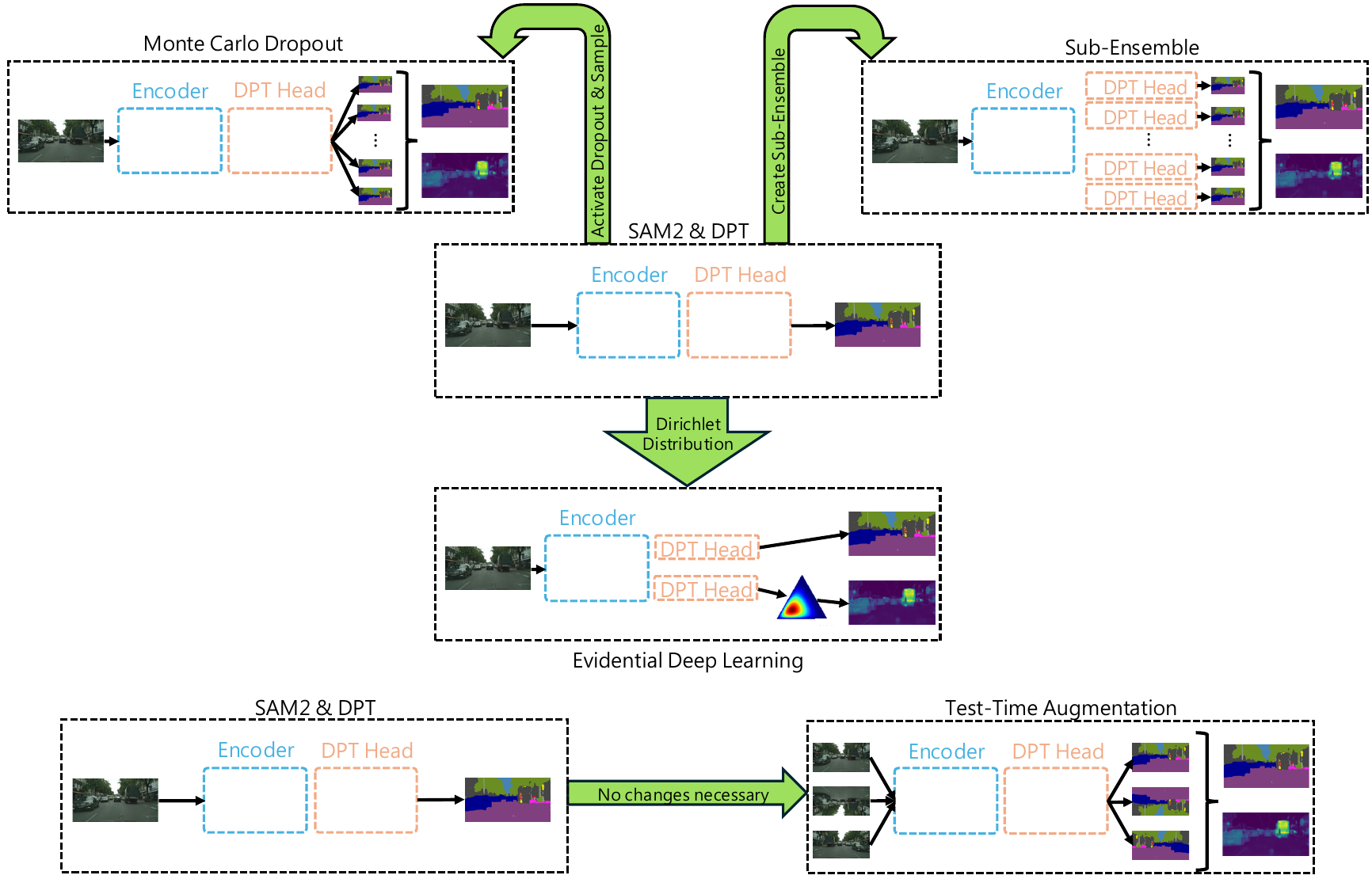}
	\caption{A schematic overview of how we combined the four different uncertainty quantification techniques with our model.}
\label{fig: method}
\end{center}
\end{figure*}

Our central research question is how to make segmentation foundation models not only accurate but also reliable in real-world deployment where well-calibrated uncertainties are essential. Following \citep{gawlikowski2023survey}, existing UQ approaches can be broadly grouped into test-time augmentation, Bayesian methods, ensembles, and deterministic methods. To cover this wide methodological spectrum, we evaluate one representative technique for each group: MCD, DSE, EDL, and TTA. While Fig.~\ref{fig: method} provides a schematic overview of how we fused our model architecture with all four UQ approaches, the following will provide a more detailed description of each.

\textbf{Monte Carlo Dropout.} MCD estimates predictive uncertainty through stochastic sampling with dropout layers. Following \citet{gal2016dropout}, dropout can be interpreted as a variational approximation to Bayesian neural networks, where the posterior distribution over weights is intractable and thus approximated by a Bernoulli distribution. By keeping dropout active during inference, each forward pass yields a slightly different softmax output $p(\hat{y}_t)$. 

To compute the predictive mean $\hat{\mu}$, we average across all $T$ stochastic samples:
\begin{equation}\label{eq: mean}
    \hat{\mu} = \frac{1}{T}\sum^T_{t=1}p(\hat{y}_t)\enspace.
\end{equation}
The corresponding uncertainty can be quantified by the variance as
\begin{equation}\label{eq: var}
    \hat{\sigma}^2 = \frac{1}{T-1}\sum^T_{t=1}(p(\hat{y}_t)-\hat{\mu})^2\enspace.
\end{equation}
Besides the standard deviation $\hat{\sigma}$, the predictive entropy can also be computed as a complete measure of the predictive uncertainty, which is composed of a combination of the aleatoric and epistemic uncertainty \citep{wolf2025calibration}:
\begin{equation}\label{eq: entropy}
    H(\hat{\mu}) = -\sum^C_{c=1}\hat{\mu}_c \log(\hat{\mu}_c)\enspace.
\end{equation} 

\textbf{Deep Sub-Ensemble.} Deep Ensembles \citep{lakshminarayanan2017simple} are considered a gold standard for UQ \citep{ovadia2019can,gustafsson2020evaluating,landgraf2025comparative}, as they capture variability across independently trained models. However, their computational cost scales linearly with the number of ensemble members, limiting their practicality in large-scale vision tasks. DSEs approximate the benefits of DEs at reduced cost by sharing most of the architecture and varying only a subset of layers close to the output head \citep{valdenegro2023sub}.

In our setup, the SAM2 encoder is shared across all ensemble members, while multiple DPT decoders are initialized independently. During training, only one decoder head is optimized per mini-batch, while the others are frozen. This cycling strategy increases diversity across decoders while keeping training efficient. At inference, each decoder in general produces a different softmax prediction $p(\hat{y}_t)$ based on the same encoder features. As in MCD, the predictive mean and predictive uncertainty are computed across all $T$ decoder heads (see Eqs. \ref{eq: mean} - \ref{eq: entropy}). 

\textbf{Evidential Deep Learning.} EDL directly models predictive uncertainty by parameterizing a Dirichlet distribution over class probabilities \citep{amini2020deep}. Instead of outputting a single categorical distribution via softmax, the network produces non-negative evidence values $e_c$ for each class $c$. These are mapped to the Dirichlet parameters
\begin{equation}
    \alpha_c = e_c + 1\enspace.
\end{equation}
The Dirichlet strength is then defined as
\begin{equation}
    S = \sum^C_{c=1} \alpha_c\enspace,
\end{equation}
where $C$ is the number of classes. From here, the belief masses $b_c$ and the total uncertainty $u$ can be obtained as
\begin{equation}
    b_c = \frac{e_c}{S}\enspace, \quad u = \frac{C}{S}\enspace.
\end{equation}
The expected class probabilities follow as
\begin{equation}
    p(\hat{y_c}) = \frac{\alpha_c}{S}\enspace.
\end{equation}
This representation enables the model to express both confident assignments -- with large $\alpha_c$ values for one class -- and uncertain states -- with low, evenly distributed $\alpha_c$ values across classes.

Training is performed with an evidential loss function based on the mean squared error, which, for a single sample $i$, can be calculated as
\begin{equation}
    \mathcal{L}_i(\Theta) = \sum_{c=1}^C \left( y_{ik} - p(\hat{y}_{ic}) \right)^2
	+ \frac{p(\hat{y}_{ic})(1 - p(\hat{y}_{ic})}{S_i + 1}\enspace,
\end{equation}
where the first term penalizes prediction errors and the second term incorporates the variance of the predictive distribution, encouraging the model to calibrate its uncertainty appropriately. To prevent overconfidence on misclassified samples, a Kullback-Leibler (KL) divergence regularization is added to obtain the final loss function
\begin{equation}
    \mathcal{L}(\Theta) = \sum^N_{i=1} \mathcal{L}_i(\Theta) + \lambda_t \sum^N_{i=1}\text{KL}[\text{D}(p(\hat{y}_i)|\tilde{a}_i) || \text{D}(p(\hat{y}_i)|1)]\enspace,
\end{equation}
where $\lambda_t$ is an annealing coefficient, $t$ is the current training epoch, $\text{D}(p(\hat{y}_i)|1)$ is the uniform Dirichlet distribution, and lastly $\tilde{\alpha}_i = y_i + (1 - y_i) \odot \alpha_i$ is the Dirichlet parameters after removal of the non-misleading evidence from predicted parameters $\alpha_i$. More details on this can be found in \citet{sensoy2018evidential}.

To stabilize training with our SAM2–DPT hybrid, we employ a two-head architecture, as shown by Fig.~\ref{fig: method}: One head predicts segmentation logits, while the second head outputs evidential uncertainty parameters. Empirically, this design significantly improved segmentation performance compared to a single shared head, where segmentation accuracy degraded drastically.

\textbf{Test-Time Augmentation.} TTA is applied only during inference to estimate the predictive uncertainty. Consequently, it is applied without modifying the model parameters or the training protocol. During inference, we generate T augmented variants of each input image and feed both the original and its augmented versions into the network for prediction. Akin to MCD and DSE, the final prediction is obtained by the mean and the corresponding uncertainty is quantified by the variance/standard deviation or predictive entropy (cf. Eqs.~\ref{eq: mean}--\ref{eq: entropy}).

\section{Experimental Setup}\label{EXPERIMENTAL_SETUP}
This section details the unified training configuration, datasets, and augmentations used in all experiments to ensure fair comparisons across UQ methods and robust evaluation under diverse conditions. 

\subsection{Training Configuration}
All models are trained for 100~epochs, which was empirically found to be sufficient for convergence. We use the AdamW optimizer~\citep{loshchilov2019decoupledweightdecayregularization} with a base learning rate of \(3\times10^{-5}\) for both datasets and a polynomial decay scheduler. As the SAM2 encoder is pre-trained while the decoder is newly initialized, we apply a learning-rate multiplier of 10 to the decoder. The loss function is the standard cross-entropy loss (cf. Eq.~\ref{eq: ce}). Due to GPU memory constraints, we set the batch size to 8 for all experiments. The initial weight decay is set to \(1\times10^{-2}\) for Cityscapes and \(1\times10^{-4}\) for NYUv2. All experiments are conducted on a single NVIDIA A100 GPU with 40\,GB VRAM.

\subsection{Datasets}
We evaluate on Cityscapes~\citep{cordts2016cityscapes} and NYUv2~\citep{silberman2012indoor}. Cityscapes contains high-resolution street scenes (2048\(\times\)1024\,px). Following standard practice, we use the 2,875 training and 500 validation images with 19 semantic classes and 1 void label, which is ignored during training. Additionally, we employ the most challenging options of the synthetic Rainy-Cityscapes~\citep{hu2019depth} and Foggy-Cityscapes~\citep{sakaridis2018semantic} variants for out-of-domain evaluation.

NYUv2 comprises 1,449 RGB-D indoor images captured with a Microsoft Kinect in 464 different rooms. We use the RGB images and the corresponding 40 semantic classes (grouped into 13 super-classes for visualization) split into 795 training and 654 test images at 640\(\times\)480\,px resolution.

The four datasets jointly allow testing under both outdoor and indoor conditions as well as in-domain and out-of-domain settings.

\subsection{Data Augmentations}
Regardless of the model, we apply random scaling with a factor between 0.5 and 2.0, random cropping with a crop size of 768\(\times\)768\,px on Cityscapes and 480\(\times\)640\,px on NYUv2, and random horizontal flipping with a flip chance of 50 \% during training as data augmentations.

\subsection{Uncertainty Quantification}
Based on rigorous evaluations, we report results based on the following UQ configurations. For TTA, we apply vertical and horizontal flipping as well as scaling during inference to generate samples to compute our predictive mean and uncertainty. We note that vertical flipping can alter spatial priors and may therefore lead to an overestimation of uncertainty in some cases -- we nevertheless found the inclusion of vertical flipping to work best. MCD uses the already existing dropout layers in the model architecture with a dropout rate of 20\% and we sample ten times during test-time. Similarly, we employ ten DPT heads for DSE.

As the predictive entropy delivered slightly better results in our evaluations than the predictive variance, we only report uncertainty metrics based on the former for all models, including the baseline.

\subsection{Metrics}
In terms of metrics, we report the mean Intersection over Union (mIoU), the Expected Calibration Error (ECE) \citep{naeini2015obtaining, wolf2025calibration}, and two uncertainty quality metrics from \citet{mukhoti2018evaluating}: 
\begin{enumerate}
    \itemsep0em
    \item \textbf{p(acc.$|$cer.):} The probability that the model is accurate on its output given that the uncertainty is below a specific threshold.
    \item \textbf{p(unc.$|$ina.):} The probability that the uncertainty of the model exceeds a specified threshold given that the prediction is inaccurate.
\end{enumerate}
Both metrics are expected to return high values for high-quality uncertainties. Based on our own empiric results and prior work from \citet{landgraf2025comparative}, we opt for the median uncertainty of any given image as the uncertainty threshold. 

\section{Experiments}\label{EXPERIMENTS}
The following section lays out numerous quantitative as well as qualitative results, encompassing not only in-domain datasets but also out-of-domain settings for a comprehensive set of experiments.  

\subsection{Quantitative Results}
\textbf{Comparison with SOTA.} Tables~\ref{tab: cityscapes} and~\ref{tab: nyu} report segmentation performance on Cityscapes and NYUv2, respectively. Without additional architectural changes or complex training strategies, our simple SAM2-DPT baseline achieves results comparable to recent state-of-the-art methods, demonstrating that fine-tuning a foundation model with a lightweight DPT head can already yield competitive performance across both datasets. For all subsequent experiments, we report results using the SAM2-Tiny backbone due to computational constraints; however, we empirically verified that the observed trends are consistent with larger backbone variants.

\begin{table}[t]
    \setlength\extrarowheight{1mm}
    \centering
    \begin{tabular}{l|c}
        \textit{Cityscapes} & mIoU $\uparrow$ \\
        \hline \hline
        DeepLabV3+ \citep{chen2018encoder} & 79.6\% \\
        U-Net++ \citep{ronneberger2015u} & 75.5\% \\
        SegFormer-B0 \citep{xie2021segformer} & 76.2\% \\
        SegFormer-B5 \citep{xie2021segformer} & 82.4\% \\
        Mask2Former \citep{cheng2022masked} & 83.3\% \\
        \hline
        SAM2-DPT [\textit{tiny}] (ours) & 76.6\% \\
        SAM2-DPT [\textit{large}] (ours) & 82.4\% \\
    \end{tabular}
    \caption{Comparison against previous approaches on Cityscapes.}
    \label{tab: cityscapes}
\end{table}

\begin{table}[t]
    \setlength\extrarowheight{1mm}
    \centering
    \begin{tabular}{l|c}
        \textit{NYUv2} & mIoU $\uparrow$ \\
        \hline \hline
        TokenFusion \citep{wang2022multimodal} & 54.2\% \\
        Omnivore \citep{girdhar2022omnivore} & 54.0\% \\
        EMSANet \citep{seichter2022efficient} & 53.3\% \\
        SGNet \citep{chen2021spatial} & 51.0\% \\
        AsymFormer \citep{du2024asymformer} & 54.1\% \\
        \hline
        SAM2-DPT [\textit{tiny}] (ours) & 43.6\% \\
        SAM2-DPT [\textit{large}] (ours) & 55.5\% \\
    \end{tabular}
    \caption{Comparison against previous approaches on NYUv2.}
    \label{tab: nyu}
\end{table}

\textbf{In-Domain Evaluation.} Table~\ref{tab: id_evaluation} shows in-domain results on Cityscapes and NYUv2. On Cityscapes, the baseline model achieves competitive mIoU, strong calibration, and offers the fastest inference time. At the same time, it offers the second worst results in terms of uncertainty quality -- only undermined by EDL, which performs worst across all metrics. DSE attains the best calibration and uncertainty quality but at the cost of a slight mIoU reduction and substantially slower inference. 

On NYUv2, the baseline achieves the highest segmentation performance and, unexpectedly, the best uncertainty quality, but shows the poorest calibration -- though these results should be interpreted cautiously given the overall low segmentation scores. EDL again provides the second-fastest inference time but performs poorly across all other metrics. TTA delivers the best calibration, while DSE follows the baseline closely in uncertainty quality.  

\begin{table*}[h]
    \centering
    \setlength{\tabcolsep}{7pt} 
    \begin{tabular}{l|c|c|c|c|c}
        \textit{in-Domain} & mIoU [\%] $\uparrow$ 
        & ECE [\%] $\downarrow$ 
        & $p(\text{acc.}|\text{cer.})$ [\%] $\uparrow$ 
        & $p(\text{unc.}|\text{ina.})$ [\%] $\uparrow$ 
        & Inference Time [ms] $\downarrow$ \\
        \midrule \midrule
        \multicolumn{6}{c}{\textbf{Cityscapes}} \\
        \midrule \midrule
        Baseline   & 76.55 & 0.95 & 92.61 & 76.18 & \textbf{58.57} \\ 
        TTA           & \textbf{76.84} & 1.76 & 94.29 & 81.32 & 202.50 \\
        MCD    & 76.10 & 1.30 & 93.52 & 80.18 & 656.07 \\
        DSE & 74.44 & \textbf{0.65} & \textbf{96.91} & \textbf{91.74} & 315.46 \\
        EDL           & 70.84 & 4.97 & 84.40 & 51.14 & 73.40 \\
        \midrule \midrule
        \multicolumn{6}{c}{\textbf{NYUv2}} \\
        \midrule \midrule
        Baseline   & \textbf{43.62} & 16.80 & \textbf{82.08} & \textbf{81.26} & \textbf{9.49} \\ 
        TTA           & 43.61 & \textbf{3.01}  & 79.53 & 77.09 & 34.52 \\
        MCD    & 38.15 & 13.15  & 78.12 & 77.26 & 124.75 \\
        DSE & 39.10 & 11.21 & 80.36 & 80.00 & 55.40 \\
        EDL           & 37.94 & 7.13  & 76.01 & 74.50 & 11.15 \\
    \end{tabular}
    \caption{Quantitative in-domain evaluation on Cityscapes and NYUv2 datasets using SAM2-DPT.}
    \label{tab: id_evaluation}
\end{table*}

\textbf{Out-of-Domain Evaluation.} Table~\ref{tab: ood_evaluation} shows out-of-domain results on Rainy- and Foggy-Cityscapes. On both datasets, segmentation performance drops markedly compared to the in-domain experiments, highlighting the vulnerability of deep learning models to domain shifts.

On Rainy-Cityscapes, TTA achieves the highest mIoU, the best calibration, and also improves the uncertainty quality compared to the baseline. MCD yields slightly higher mIoU than TTA but a substantially worse calibration. DSE performs comparably to the baseline with worse calibration but higher p(unc.$|$ina.). EDL again shows the weakest results across all metrics.

For Foggy-Cityscapes, TTA gives the best mIoU at the cost of notably worse calibration and uncertainty quality, whereas DSE attains the best uncertainty scores, especially p(unc.$|$ina.). MCD remains close to the baseline with moderate improvements in mIoU but worse calibration. EDL again performs worst on all metrics. 

\begin{table*}[h]
    \centering
    \setlength{\tabcolsep}{14pt}
    \begin{tabular}{l|c|c|c|c}
        \textit{Out-of-Domain} & mIoU [\%] $\uparrow$ 
        & ECE [\%] $\downarrow$ 
        & $p(\text{acc.}|\text{cer.})$ [\%] $\uparrow$ 
        & $p(\text{unc.}|\text{ina.})$ [\%] $\uparrow$  \\
        \midrule \midrule
        \multicolumn{5}{c}{\textbf{Rainy-Cityscapes}} \\
        \midrule \midrule
        Baseline   & 46.72 & 5.09 & 92.66 & 80.92 \\ 
        TTA        & 48.99 & \textbf{0.84} & \textbf{93.40} & \textbf{83.85} \\
        MCD        & \textbf{49.37} & 5.33 & 92.29 & 78.28 \\
        DSE        & 45.71 & 7.07 & 92.65 & 83.21 \\
        EDL        & 40.91 & 12.24 & 89.68 & 75.12 \\
        \midrule \midrule
        \multicolumn{5}{c}{\textbf{Foggy-Cityscapes}} \\
        \midrule \midrule
        Baseline      & 53.96 & \textbf{3.09} & 89.54 & 80.11 \\ 
        TTA           & \textbf{56.53} & 7.46  & 88.48 & 77.43 \\
        MCD           & 54.49 & 5.67  & 90.39 & 81.44 \\
        DSE           & 51.14 & 4.82 & \textbf{93.40} & \textbf{89.49} \\
        EDL           & 50.97 & 14.26  & 83.64 & 67.09 \\
    \end{tabular}
    \caption{Quantitative out-of-domain evaluation on Rainy- and Foggy-Cityscapes using SAM2-DPT.}
    \label{tab: ood_evaluation}
\end{table*}

\subsection{Qualitative Results}

\begin{figure*}[t!]
\begin{center}
		\includegraphics[width=1.0\linewidth]{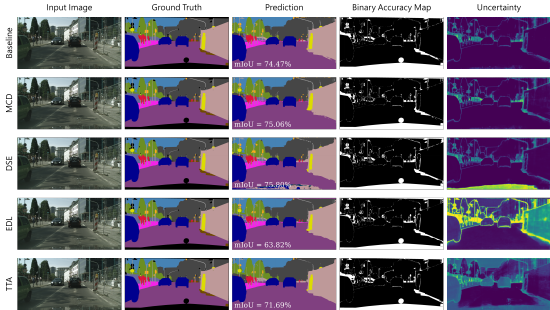}
	\caption{Qualitative in-domain evaluation on Cityscapes dataset using SAM2-DPT.}
\label{fig: qualitative_evaluation}
\end{center}
\end{figure*}

Fig.~\ref{fig: qualitative_evaluation} shows in-domain qualitative results on Cityscapes. The baseline, MCD, and DSE produce the most accurate segmentation predictions in terms of mIoU, whereas TTA performs slightly worse and EDL exhibits the worst results. The corresponding binary accuracy maps reveal largely similar falsely segmented regions across all methods. Regarding the uncertainty, TTA and EDL show a higher and spatially more widespread uncertainty, while the baseline, MCD and DSE display more localized and lower levels of uncertainty. Notably, DSE captures the erroneous regions most effectively, as its uncertainties align well with the binary error maps, indicating a strong correspondence between prediction errors and uncertainty estimates.

\onecolumn
\clearpage
\twocolumn

\subsection{Discussion}\label{DISCUSSION}
Our results reveal several important patterns regarding the critical synthesis of UQ and foundation models for segmentation. Although the SAM2–DPT baseline achieves competitive segmentation accuracy with minimal architectural changes, its calibration and uncertainty quality are insufficient and inconsistent across domains. This confirms that high predictive performance does not automatically translate to high reliability.

Additionally, we found clear trade-offs w.r.t. the UQ strategies. While DSE offers high uncertainty quality, it comes at a significant cost of inference speed. TTA improves upon the baseline in most cases but also suffers from high computation cost during inference. MCD, despite the highest inference times, often underperforms even the baseline, and EDL yields the least promising results. These observations highlight that there is no single best method and that the choice of UQ strategy highly depends on the deployment context.

\section{Conclusion}\label{CONCLUSION}
We presented the first systematic evaluation of multiple UQ methods applied to a SAM2-based foundation model for semantic segmentation. Our experiments across Cityscapes, NYUv2 and two out-of-domain settings show that competitive segmentation accuracy is easily attainable with a lightweight decoder, but reliability and efficiency differ markedly between UQ strategies. It is clear that there is no one-size-fits-all solution yet. This underscores the importance of future work not only focusing on predictive performance but also reliability and efficiency. 

{
\begin{spacing}{0.94}
    \normalsize
    \bibliography{ISPRSguidelines_authors} 
\end{spacing}
}

\end{document}